\documentclass[lettersize,journal]{IEEEtran}



\usepackage{amsmath,amsfonts}
\usepackage{algorithm, algorithmic}
\usepackage{graphicx}
\usepackage[caption=false,font=normalsize,labelfont=sf,textfont=sf]{subfig}
\usepackage{booktabs,array,stfloats,multirow}
\usepackage[table]{xcolor}
\usepackage{cite,hyperref}
\hypersetup{hidelinks}

\def\BibTeX{{\rm B\kern-.05em{\sc i\kern-.025em b}\kern-.08em
    T\kern-.1667em\lower.7ex\hbox{E}\kern-.125emX}}
\usepackage{balance}

\begin{document}
\title{Detect, Explain, Escalate: Sustainable Dialogue Breakdown Management for LLM Agents}
\author{Abdellah Ghassel, Xianzhi Li, and Xiaodan Zhu,~\IEEEmembership{Member,~IEEE}
\thanks{Manuscript received June 1, 2025; revised November 27, 2025; accepted December 31, 2025. The associate editor coordinating the review of this manuscript and approving it for publication was Prof. Mark Hasegawa-Johnson. \textit{(Corresponding author: Abdellah Ghassel.)}}
\thanks{The authors are with the Department of Electrical and Computer Engineering and Ingenuity Labs Research Institute, Queen's University, Kingston, ON K7L 3N6, Canada (e-mail: abdellah.ghassel@queensu.ca; li.xianzhi@queensu.ca; xiaodan.zhu@queensu.ca).}
}

\markboth{Journal of IEEE Transactions on Audio, Speech and Language Processing, Vol. 00, No. 0, Month 2025}
{Ghassel \MakeLowercase{\textit{et al.}}: Detect, Explain, Escalate: Sustainable Dialogue Breakdown Management for LLM Agents}

\maketitle

\begin{abstract}
Large Language Models (LLMs) have demonstrated substantial capabilities in conversational AI applications, yet their susceptibility to dialogue breakdowns poses significant challenges to deployment reliability and user trust. This paper introduces a ``Detect, Explain, Escalate'' framework to manage dialogue breakdowns in LLM-powered agents, emphasizing resource-efficient operation. Our approach integrates two key strategies: (1) We fine-tune a compact 8B-parameter model, augmented with teacher-generated reasoning traces, which serves as an efficient real-time breakdown \emph{detector} and \emph{explainer}. This model demonstrates robust classification and calibration on English and Japanese dialogues, and generalizes to the BETOLD dataset, improving accuracy by 7\% over its baseline. (2) We systematically evaluate frontier LLMs using advanced prompting (few-shot, chain-of-thought, analogical reasoning) for high-fidelity breakdown assessment. These are integrated into an ``escalation'' architecture where our efficient detector defers to larger models only when necessary, substantially reducing operational costs and computational overhead. Our fine-tuned model and prompting strategies achieve state-of-the-art performance on DBDC5 and strong results on BETOLD, outperforming specialized classifiers on DBDC5 and narrowing the performance gap to larger proprietary models. The proposed monitor--escalate pipeline reduces inference costs by 54\%, providing a cost-effective and interpretable solution for robust conversational AI in high-impact domains. Code and models are publicly available.\footnote{\url{https://github.com/aghassel/dee-dialogue-agents}}
\end{abstract} 

\begin{IEEEkeywords}
large language models, conversational artificial intelligence, dialogue systems, dialogue breakdown detection, human--computer interaction, model calibration
\end{IEEEkeywords}

\section{Introduction}
Conversational artificial intelligence (AI) has experienced substantial growth driven by advances in large language models (LLMs) such as ChatGPT~\cite{openai_gpt-4_2023}, Claude~\cite{ClaudeModelCard}, and Llama~\cite{llama}. ChatGPT's rapid adoption, reaching 100 million users within two months~\cite{Yatoo2023ChatGPT,Khosravi2023Chatbots}, demonstrates the widespread interest in LLM-based conversational systems and their potential to transform human-computer interaction. These models are now embedded in applications spanning customer service, content creation, marketing, and education, enabling new forms of human–AI interaction~\cite{Currie2023A,Gabriella2024The,Arviani2023ChatGPT}.

\begin{table}[!t]
\centering
\caption{Sample dialogue illustrating contradictory turns.}
\label{tab:sample-dialogue}
\begin{minipage}{0.97\linewidth}   
\begin{tabular}{@{}p{\linewidth}@{}}
\toprule
1.\,\textbf{Assistant}: It’s nice to go \underline{shopping alone}.\\
2.\,\textbf{User}: I agree. That’s nice.\\
3.\,\textbf{Assistant}: Shopping takes time.\\
4.\,\textbf{User}: Window shopping is also fun.\\
\addlinespace
\multicolumn{1}{c}{\textit{Determine if the next utterance causes a breakdown:}}\\
\addlinespace
5.\,\textbf{Assistant}: It’s fun to go \underline{shopping with somebody}.\\
\bottomrule
\end{tabular}
\end{minipage}
\end{table}

Despite these advancements, the accelerated integration of LLMs into critical domains has revealed significant challenges in dialogue breakdown detection and mitigation, which directly impact user trust and conversational effectiveness~\cite{Ghassel2024Are,Fan2022Building,candello_recovering_2018}. Dialogue breakdowns typically manifest as lapses in conversational coherence, leading to irrelevant, contradictory, or incoherent exchanges that negatively impact human-AI interactions~\cite{higashinaka_dialogue_2016,Ashktorab}. Table \ref{tab:sample-dialogue} illustrates a common breakdown pattern, where the assistant’s utterances contradict one another.

As LLM-based systems are deployed in high-stakes environments~\cite{Currie2023A,Gabriella2024The,Arviani2023ChatGPT}, robust handling of such breakdowns becomes increasingly important. The tendency of LLMs to produce overly confident yet inaccurate or hallucinated outputs further complicates safe deployment~\cite{xiong_can_2023,zhou_larger_2024}. There is therefore a pressing need for methods that not only detect dialogue breakdowns reliably but also support mitigation strategies that preserve conversational quality and user trust.

Traditionally, dialogue breakdown detection has been approached using specialized classifiers trained on labelled datasets such as the Dialogue Breakdown Detection Challenge (DBDC)~\cite{higashinaka_dialogue_2016,tsunomori-etal-2018-evaluating}. While these models achieve strong benchmark performance, their generalization to diverse, real-world contexts remains limited. At the same time, recent work shows that general-purpose LLMs still lag behind human performance for nuanced conversational behavior, despite their broad coverage and world knowledge~\cite{finch_leveraging_2023,finch_dont_2023}. This gap indicates that neither specialized classifiers nor generalist LLMs alone provide a complete solution.

In this paper, we propose a framework that combines the strengths of both approaches. We leverage the reasoning capabilities of generalist LLMs through supervised fine-tuning and structured prompting, and we embed these capabilities within a resource-aware deployment architecture.

First, we fine-tune a parameter-efficient \texttt{Llama-3.1 8B} model~\cite{llama} on both English and Japanese tracks of the DBDC5 dataset~\cite{Higashinaka2020DBDC5}, augmenting supervision with synthetic reasoning trajectories generated by a larger teacher model, \texttt{Llama-3.3 70B}. These distilled reasoning traces are used during training to shape the student model’s decision process and to support interpretable justifications at inference time. We then evaluate the model’s generalization on the BETOLD dataset~\cite{terragni2022_BETOLD}, which focuses on task-oriented dialogue breakdowns in realistic customer-service settings.

Second, we conduct a comparative analysis of both closed-source frontier models (OpenAI~\cite{openai_gpt-4_2023}, Anthropic~\cite{ClaudeModelCard}) and open-source alternatives (Meta~\cite{llama}, Mistral~\cite{jiang_mixtral_2024}, DeepSeek~\cite{deepseek}) on dialogue breakdown detection. We explore and propose novel prompting schemes, including few-shot learning~\cite{brown_language_2020,palm2023}, chain-of-thought prompting~\cite{wei_chain--thought_2023}, and analogical reasoning with curriculum learning~\cite{analogical,xu-etal-2020-curriculum,Ryu2024CurricuLLM:}, to elicit more systematic reasoning while regulating token costs.

Third, we integrate these components into a hierarchical, cost-aware deployment architecture. A fine-tuned \texttt{Llama-3.1 8B} monitor provides fast, per-turn breakdown detection and explanation and selectively escalates to more capable models (\texttt{GPT-4}, \texttt{DeepSeek-R1}, \texttt{Claude-3.5 Sonnet}, or \texttt{Llama-3.1 405B}) only when its confidence indicates elevated risk. This architecture yields substantial cost and energy savings without sacrificing accuracy.

We summarize our contributions as follows:
\begin{itemize}
\item We conduct, to our knowledge, the first comprehensive comparative analysis of open-source and frontier closed-source LLMs, establishing new benchmarks on dialogue breakdown detection tasks.
\item We jointly evaluate accuracy and calibration, revealing important differences in reliability and overconfidence across models, prompting strategies, and languages.
\item We propose a real-time deployment architecture that significantly reduces operational costs and optimizes resource usage by selectively invoking large-scale models.
\item We demonstrate that monitor-generated explanations can effectively guide a superior model to repair problematic responses; conditioning the superior model on these justifications resolves 97\% of sampled breakdowns.
\end{itemize}

\section{Related Work}
Ensuring robustness in conversational AI systems, particularly for dialogue breakdown detection and mitigation, is a central research challenge. Dialogue breakdowns occur when coherence or relevance is disrupted, hindering the smooth progression of a conversation and reducing user satisfaction~\cite{higashinaka_dialogue_2016, Ashktorab}. Practical conversational agents must therefore both detect breakdowns and employ recovery strategies to maintain engagement and trust~\cite{mctear, skantze-2017-towards}.

\subsection{Specialized Models for Dialogue Breakdown Detection}
Research into dialogue breakdown detection has produced specialized classifiers that perform strongly on benchmarks such as DBDC5 (see Section \ref{dialogue-breakdown-datasets})~\cite{higashinaka_dialogue_2016}. Leading approaches rely on fine-tuned transformer encoders~\cite{attentionIsAll}. For example, the best-performing model on the DBDC5 English track, \texttt{BERT+SSMBA}~\cite{Devlin2019BERT, ng_improving_2020}, augments a BERT-based classifier with unlabelled dialogue data through extended pre-training on dialogue-rich corpora (e.g., Reddit) and Self-Supervised Manifold-Based Data Augmentation (SSMBA)~\cite{ng_ssmba_2020}. This design improves robustness by exploiting large quantities of unlabelled conversational data.

Semi-supervised approaches further advance the state of the art. The \texttt{S2T2} model employs a dual-teacher paradigm that leverages both labelled and unlabelled dialogues~\cite{lin_semi-supervised_2022}. One teacher is trained on high-quality labelled data, while the other is trained on masked dialogue variants. A student model is jointly guided by both teachers. \texttt{S2T2} reaches new state-of-the-art performance on DBDC5 using \texttt{RoBERTa-large} on the English track and \texttt{XLM-R-large} with a context-matching mechanism on the Japanese track~\cite{liu2019roberta, conneau2020unsupervised, lin_semi-supervised_2022}.

More recently, general-purpose LLMs have been evaluated for dialogue breakdown detection. Finch et al.~\cite{finch_leveraging_2023, finch_dont_2023} studied ChatGPT’s behaviour across nine categories in the ABC-Eval dataset. While ChatGPT outperforms specialized models on some categories (e.g., empathetic behaviour), it still falls short of human performance on many dialogue tasks. This suggests that current large models, despite their sophisticated capabilities, remain unreliable for fine-grained breakdown detection. To our knowledge, prior work has not systematically evaluated multiple LLM families on dialogue breakdown detection and remediation.

\subsection{Current State of Conversational Agents}
Beyond detection, modern conversational agents increasingly incorporate mechanisms to mitigate or recover from breakdowns. In the DBDC5 recovery track, for instance, systems respond to detected breakdowns by asking clarifying questions or providing corrections~\cite{higashinaka_dialogue_2016}. Industrial systems such as ChatGPT and Claude are trained with alignment techniques like reinforcement learning from human feedback (RLHF) to reduce toxic, incoherent, or nonsensical outputs~\cite{openai_gpt-4_2023, ClaudeModelCard}. Claude-2, for example, uses internal debate and self-critique during training to identify and minimize reasoning flaws~\cite{claude2}. These mechanisms reduce the incidence of breakdowns but do not eliminate them, especially under distribution shift or adversarial inputs.

\subsection{Techniques in Conversational AI}
A variety of techniques have been proposed to improve the reliability of LLM-based conversational agents:

\noindent \textbf{Analogical Reasoning}.
Analogical prompting encourages a model to draw on relevant past examples by analogy when solving a new problem. Instead of relying on hand-crafted exemplars, the model is prompted to generate analogous examples in context and then solve the target query~\cite{analogical, webb_emergent_2023}. This approach can improve reasoning by creating problem-specific exemplars without additional labelling.

\noindent \textbf{Chain-of-Thought Reasoning}.
Chain-of-thought (CoT) prompting instructs the model to produce intermediate reasoning steps before giving a final answer~\cite{zs-cot, fs-cot, wei_chain--thought_2023}. CoT has been shown to improve performance on arithmetic, logical, and commonsense reasoning tasks. For dialogue agents, CoT can be applied internally to reason about user intent, dialogue history, and knowledge bases, thereby reducing non sequiturs and hallucinations.

\noindent \textbf{Zero-Shot and Few-Shot Learning}. Unlike traditional dialogue systems requiring extensive task-specific training, LLMs excel at in-context learning, adapting to new tasks given instructions and a small number of examples. In few-shot learning, a prompt may include example dialogues or question–answer pairs, and the model generalizes the pattern to new inputs. \texttt{GPT-3} demonstrated that large models can function as few-shot learners without task-specific fine-tuning~\cite{brown_language_2020}.

\subsection{Hierarchical Architectures}
Hierarchical or multi-tiered architectures are a common strategy for balancing performance and cost in AI systems~\cite{chen2024frugalgpt, yue2024large}. In the context of LLMs, such architectures typically use smaller, faster models for routine inference and reserve larger, more capable models for complex or high-stakes cases~\cite{10.5555/3692070.3693614, 10.5555/3586589.3586709}. This can be viewed as a system-level “mixture of experts”~\cite{2017olln}, where routing decisions balance accuracy and computational expense. Our work extends this paradigm for real-time dialogue management, positioning a lightweight monitor in front of more capable models to create a practical and sustainable framework for conversational agents.

\section{Problem Definition}
Dialogue breakdown refers to the deterioration of coherence, relevance, or conversational fluency between a user and a conversational agent~\cite{higashinaka_dialogue_2016, skantze-2017-towards}. Breakdowns may appear as irrelevant responses, misunderstandings, contradictions, or incoherent interactions that disrupt the dialogue and erode user trust~\cite{Fan2022Building, candello_recovering_2018}. As LLMs from the OpenAI~\cite{openai_gpt-4_2023}, Claude~\cite{ClaudeModelCard}, and Llama~\cite{llama} families are deployed in diverse conversational tasks, detecting and mitigating such breakdowns becomes critical. This is exacerbated by LLMs’ propensity to produce confident but incorrect or hallucinated content~\cite{xiong_can_2023,zhou_larger_2024}.

We consider a multi-turn dialogue sequence \(D\) between a user (U) and a conversational agent (A). At each turn \(i\), the user produces an utterance \(u_i\), and the agent responds with \(s_i\). The dialogue can be written as:
\[
D = \bigl(u_1, s_1,\; u_2, s_2,\;\dots,\; u_n, s_n\bigr).
\]
We aim to detect, at each agent utterance \(s_i\), whether the conversation has experienced a breakdown in coherence, relevance, or consistency.

\subsection{Utterance-Level Breakdown Detection}

Let \(\mathcal{H}_i\) denote the contextual history available just before the agent produces its \(i\)-th response:
\[
\mathcal{H}_i = \bigl(u_1, s_1, \dots, u_{i-1}, s_{i-1}, u_i\bigr).
\]
We define a classification function \(f\) that, given \(\mathcal{H}_i\) and the agent’s latest utterance \(s_i\):
\[
 f: \bigl(\mathcal{H}_i,\; s_i\bigr) \mapsto (\hat{b}_i,\;\hat{c}_i,\;\hat{j}_i),
\]
where:
\begin{itemize}
    \item \(\hat{b}_i\in\{0,1\}\) is a binary classification indicating dialogue breakdown (1) or non-breakdown (0).
    \item \(\hat{c}_i \in [0,1]\) is a confidence score representing the model's certainty about the predicted label.
    \item \(\hat{j}_i\) is a textual justification explaining the model's reasoning process.
\end{itemize}

For a dialogue \(D\) with \(n\) system turns, the detector produces:
\[
    \mathcal{O}(D) = \{(b_i, c_i, j_i)\}_{i=1}^{n}
\]
\subsection{Consolidation of Three-Class Annotations}
Following prior works~\cite{Ghassel2024Are, lin_semi-supervised_2022}, we preprocess DBDC5’s three-way labels (Breakdown (B), Possible Breakdown (PB), and Non-Breakdown (NB)) into a binary Breakdown/Non-Breakdown scheme. Ambiguous annotations (PB) are merged into the Breakdown class. This specific binarization strategy aligns with the evaluation protocol employed by the S2T2 baseline~\cite{lin_semi-supervised_2022}, ensuring that the performance comparisons reported in Section~\ref{res-and-disc} are methodologically valid. 

When multiple annotators label each system turn, let \(p_i\) be the fraction who assign B or PB to utterance \(s_i\). We define a binary label \(b_i\) by thresholding:
\[
    b_i \;=\;
    \begin{cases}
       1, & \text{if } p({b_i}\mid s_i) \,\geq\, 0.5,\\[4pt]
       0, & \text{otherwise},
    \end{cases}
\]
where \(p({b_i} \mid s_i)\) denotes the fraction of annotators labelling \(s_i\) as a breakdown.  

In deployment, the monitor must provide a binary control signal (continue vs.\ intervene) under tight latency constraints and asymmetric risk: missing a breakdown (false negative) is often more costly than a spurious intervention (false positive). Consolidating PB into the Breakdown class yields a conservative detector that treats boundary cases as risky. This intentionally increases the false-positive rate to reduce false negatives. The decision threshold \(T\) on \(\hat{c}_i\) can then be tuned to balance sensitivity and specificity for a given application.

\subsection{Conversation-Level Labelling}
In task-oriented datasets such as BETOLD~\cite{terragni2022_BETOLD}, labels are assigned at the conversation level. Each dialogue \(D\) is marked as a failure if, for example, the user hangs up or requests escalation to a human agent. We denote this label by:
\[
\mathcal{O}(D) \;=\;
\begin{cases}
1, & \text{if the conversation leads to breakdown},\\
0, & \text{otherwise}.
\end{cases}
\]
Here the model predicts a single label after observing all \((u_i, s_i)\) pairs in \(D\). Conversation-level breakdown often results from one or more local breakdown events, but the relationship is not necessarily one-to-one.

\subsection{Confidence Calibration}
An effective breakdown detector should classify accurately and calibrate its confidence well. Let \(p_i\) be the true probability of a breakdown for the \(i\)-th utterance (estimated from multiple annotators) and \(\hat{c}_i\) be the model’s predicted probability. We measure calibration quality via mean-squared error (MSE):
\[
\mathrm{MSE} \;=\; \frac{1}{N}\sum_{i=1}^{N}\Bigl(\,\hat{c}_i \;-\; p_i\Bigr)^2,
\]
where \(N\) is the total number of utterances in the test set. Lower MSE indicates that the model’s self-reported confidence aligns more closely with actual annotator distributions.

\section{Methodology}
Our approach combines supervised fine-tuning with advanced prompting strategies to achieve accurate dialogue breakdown detection in both open-domain and task-oriented settings. The framework comprises three components: (1) a compact fine-tuned model (\texttt{Llama-3.1 8B}) for efficient real-time breakdown detection, (2) prompting strategies (few-shot, chain-of-thought, and analogical reasoning) for large-scale LLMs, and (3) a multi-tier inference architecture that escalates to high-capacity models only when confidence thresholds indicate potential breakdowns.

\subsection{Supervised Fine-tuning with Reasoning Augmentation}
To support real-time monitoring, we fine-tune a smaller model on labelled breakdown data using supervised fine-tuning (SFT). We select \texttt{Llama-3.1 8B}~\cite{llama} as a balance between accuracy and computational cost. The model is fine-tuned on the DBDC5 English and Japanese tracks~\cite{higashinaka_dialogue_2016} for per-utterance breakdown labels.

Let:
\begin{itemize}
    \item \(\mathcal{D} = \{(\mathcal{H}_i, s_i, b_i)\}_{i=1}^{N}\) denote the training data, where each sample comprises a context \(\mathcal{H}_i\), agent utterance \(s_i\), and a binary label \(b_i \in \{0,1\}\).  
    \item \(T\) be a larger teacher LLM, such as \texttt{Llama-3.3 70B}, which generates synthetic reasoning traces \(r_i\) (chain-of-thought style explanations) for each sample \((\mathcal{H}_i, s_i,  b_i)\).
\end{itemize}

We augment the training inputs with these synthetic reasoning traces, forming an enriched dataset \(\mathcal{D}'\). The student model \(S\) is fine-tuned on \(\mathcal{D}'\) to (i) predict the binary breakdown label \(b_i\) by minimizing cross-entropy loss and (ii) generate a textual justification \(\hat{j}_i\). Teacher-generated traces \(r_i\) serve as targets for justification, encouraging the student to internalize structured reasoning patterns that improve both classification and interpretability. The cross-entropy loss~\cite{shannon_cel} is:
\[
    \mathcal{L}_{CE} = -\sum_{i=1}^{N}[y_i \log \hat{y}_i + (1 - y_i)\log(1 - \hat{y}_i)],
\]
where \(y_i\) is the true breakdown label and \(\hat{y}_i\) is the predicted breakdown probability. During training, the student observes both the original dialogue context \((\mathcal{H}_i, s_i)\) and the teacher’s reasoning \(r_i\), enabling distillation of richer decision rules than supervision on labels alone.

\subsection{Prompting Strategies}
While a fine-tuned compact model is suitable for continuous monitoring, larger LLMs can still be leveraged for higher-fidelity assessment under carefully designed prompts. Let \(\mathcal{G}\) denote a generalist LLM such as \texttt{GPT-4} or \texttt{DeepSeek-R1}. Given \((\mathcal{H}_i, s_i)\), we form a prompt \(\Pi(\mathcal{H}_i,s_i;\,\alpha)\) under strategy \(\alpha\in\{\mathrm{ZS}, \mathrm{FS}, \mathrm{CoT}, \mathrm{AR}, \mathrm{CL+AR},\dots\}\). The model output is parsed into a breakdown label and confidence.

\begin{figure}
    \centering
    \includegraphics[width=1.0\linewidth]{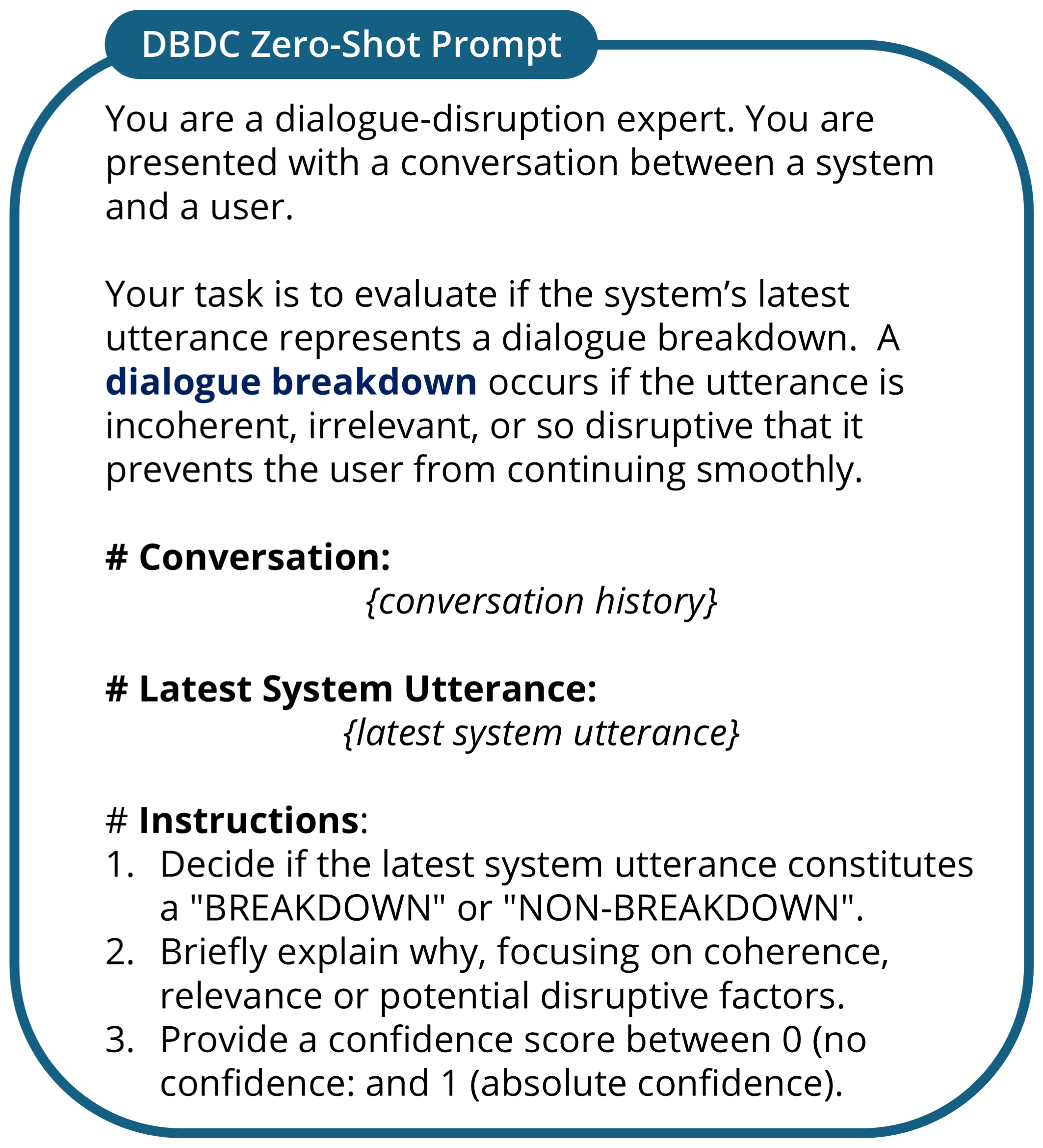}
    \caption{Sample Zero-Shot Prompt for DBDC5}
    \label{fig:sample-zs}
\end{figure}

\noindent \textbf{Zero-Shot (ZS) Prompting.}  
In the zero-shot setting, we provide a natural language task description and the current example \((\mathcal{H}_i, s_i)\). The model must infer the decision boundary from instructions alone. A sample DBDC5 zero-shot prompt is shown in Figure \ref{fig:sample-zs}.

\noindent \textbf{Chain-of-Thought (CoT).}
CoT encourages multi-step reasoning by instructing \(\mathcal{G}\) to ``think step by step'' before deciding:
\[
\Pi_\mathrm{CoT}(\mathcal{H}_i, s_i) \;=\; \bigl(\mathcal{H}_i,\,s_i,\,\text{``Let’s think step by step.''}\bigr).
\]
This often improves predictions on complex cases, though it increases token usage.

\noindent \textbf{Few-Shot (FS) Prompting.}
Few-shot prompts prepend \(k\) labelled exemplars before the test example:
\[
\Pi_\mathrm{FS}(\mathcal{H}_i, s_i;\,k) \;=\; 
\bigl(\,(\mathcal{H}_1', s_1', b_1'),\dots,(\mathcal{H}_k', s_k', b_k'),\, (\mathcal{H}_i, s_i)\bigr).
\]
We consider several variants:
\begin{itemize}
    \item \emph{2-Shot Easy (2S-Easy)}: two clear dialogues (one breakdown, one non-breakdown) with high annotator agreement (\(>80\%\)). For example, an easy non-breakdown case might be a smooth dialogue that successfully completes, and an easy breakdown case might feature an obvious user hang-up after a system error. This helps the model anchor on unambiguous prototypes.
    \item \emph{2-Shot Hard (2S-Hard)}: exemplars with moderate annotator agreement (60--70\%), representing borderline cases. These can improve calibration by exposing the model to uncertainty. One example might be a conversation with some confusion that eventually recovers (almost a breakdown, but not quite), and another might show a user mildly frustrated (not a clear-cut hang-up, but dialogue quality is low).
    \item \emph{4-Shot (4S)}: a mix of two easy and two hard exemplars, providing broader coverage at the cost of longer prompts.
\end{itemize}

For BETOLD, which lacks per-utterance rationales, we select dialogues by length (15–20 vs.\ 21–30 turns) as proxies for easier and harder cases. Since the annotators did not provide their reasoning, we generate the step-by-step reasoning field for each example using Llama-3.3 70B, given the annotators' probability distribution (treated as a confidence score) and the decision label.

\noindent \textbf{Analogical Reasoning (AR).}
In analogical prompting, we instruct \(\mathcal{G}\) to construct hypothetical dialogues that are analogous to \((\mathcal{H}_i,s_i)\) and then classify the original example. Formally,
\[
\mathcal{A}_i \;=\; \mathcal{G}\!\bigl(\Pi_\mathrm{AR}(\mathcal{H}_i,s_i)\bigr),
\]
where \(\Pi_\mathrm{AR}\) asks the model to ``recall or construct past dialogues similar to this one.'' The final decision uses both the original dialogue and generated analogies \(\mathcal{A}_i\). This self-generated context removes the need for manually crafted exemplars.

\noindent \textbf{Curriculum Learning with Analogical Reasoning (CL+AR).}
Curriculum learning organizes examples from easy to hard. We apply this idea to analogical prompting by asking the LLM to generate a sequence of analogies \(\mathcal{A}_i^{(1)}, \dots, \mathcal{A}_i^{(m)}\) that gradually increase in difficulty, culminating in the target-like scenario:
\[
\Pi_\mathrm{CL+AR}(\mathcal{H}_i,s_i) 
= \bigl(\mathcal{A}_i^{(1)},\,\mathcal{A}_i^{(2)},\,\dots,\,\mathcal{A}_i^{(m)},\,(\mathcal{H}_i,s_i)\bigr).
\]
The hope is that solving simpler analogous cases first helps the model reason more reliably about the true example.

\subsection{Deployment Architecture}
\begin{figure*}[h]
    \centering
    \includegraphics[width=\textwidth]{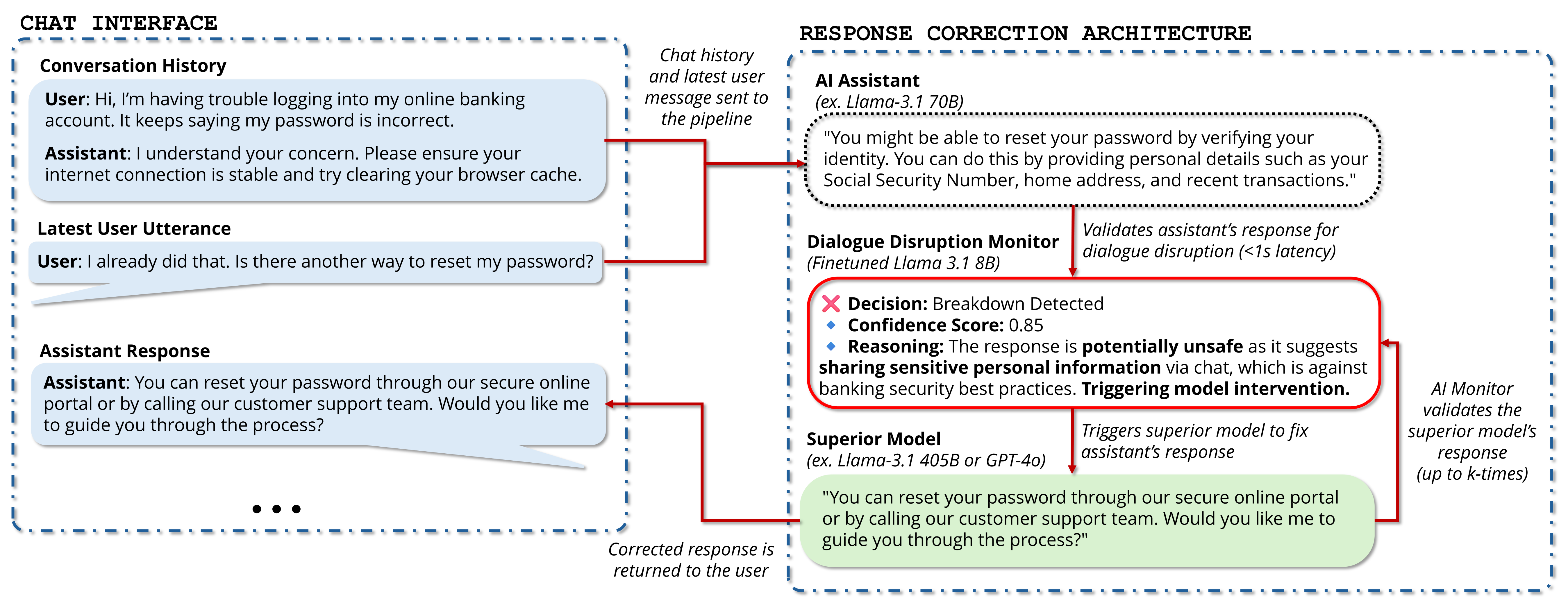}
    \caption{Real-time Response Correction Architecture. The dialogue disruption monitor intercepts potentially unsafe assistant responses, triggering a correction from a superior model before presenting the response to the user.}
    \label{fig:deployment-architecture}
\end{figure*}

While large-scale LLMs demonstrate reliable breakdown detection capabilities, their computational cost and inference latency make continuous deployment impractical for real-time applications. We therefore adopt a hierarchical architecture (Figure \ref{fig:deployment-architecture}) with three modules:

\begin{enumerate}
    \item \textbf{AI Assistant}:
   An assistant model generates a candidate response \(s_i\) given the user’s input \(u_i\) and history \(\mathcal{H}_i\):
   \[
   s_i \;=\; \mathcal{G}_{\mathrm{assistant}}\!\bigl(\mathcal{H}_i,\;u_i\bigr),
   \]
   where \(\mathcal{G}_{\mathrm{assistant}}\) may be a moderately sized LLM such as \texttt{Llama-3.3 70B}.

    \item \textbf{Dialogue Disruption Monitor}:
    Before returning \(s_i\) to the user, our fine-tuned monitor evaluates it for potential breakdowns or unsafe behaviour:
    \[
   (\hat{b}_i,\;\hat{c}_i,\;\hat{j}_i) \;=\; \mathcal{G}_{\mathrm{monitor}}\!\bigl(\mathcal{H}_i,\;s_i\bigr),
    \]
   where \(\hat{b}_i\) is the breakdown label, \(\hat{c}_i\) is a confidence score, and \(\hat{j}_i\) is an optional justification. If \(\hat{b}_i = 0\) and \(\hat{c}_i < T\), we accept \(s_i\). Otherwise, we escalate.
   
    \item \textbf{Superior Model}:
    When escalation is triggered, a more capable model revises the response:
    \[
    s_i^{*} \;=\; \mathcal{G}_{\mathrm{superior}}\!\bigl(\mathcal{H}_i, u_i, \hat{j}_i\bigr),
    \]
    producing a corrected response \(s_i^{*}\) using the history, user input, and monitor’s justification. Conditioning on \(\hat{j}_i\) (e.g., ``response is contradictory'') allows the superior model to apply targeted repairs rather than recomputing from scratch.

\end{enumerate}

In customer-service scenarios, a high-confidence breakdown prediction (\(\hat{c}_i > T\)) could additionally trigger a silent alert to a human agent. The alert includes \(\hat{j}_i\), enabling efficient human review and intervention. While the superior model could in principle act as both assistant and monitor, doing so at every turn would negate the efficiency gains of the hierarchy. The lightweight monitor is thus essential for scalable and sustainable deployment.

\section{Experiments}
We evaluate our approach on three dialogue breakdown detection benchmarks: the English and Japanese tracks of DBDC5 and the BETOLD dataset for task-oriented dialogues.

\subsection{Dialogue Breakdown Datasets}\label{dialogue-breakdown-datasets}
\noindent\textbf{Dialogue Breakdown Detection Challenge 5 (DBDC5)}.
DBDC5, introduced at the WOCHAT IWSDS 2020 workshop, is a standard benchmark for dialogue breakdown detection~\cite{Higashinaka2020DBDC5,higashinaka_dialogue_2016}. The evaluation set includes 1{,}950 English and 2{,}672 Japanese dialogues between humans and conversational agents. Each system utterance is annotated by 15–30 human annotators into three categories: Breakdown, Non-Breakdown, and Possible Breakdown. The agents span rule-based, retrieval-based, and neural generative architectures. Although these differ from modern LLM-based systems, the underlying breakdown phenomena (topic drift, contradiction, ignored context) remain salient sources of error~\cite{Huang2025Taxonomy}.

\noindent\textbf{Breakdown Expectation for Task-Oriented Long Dialogues (BETOLD)}. Introduced in 2022, BETOLD targets breakdowns in real-world, task-oriented customer-service calls~\cite{terragni2022_BETOLD}. It contains 13{,}524 human–agent phone dialogues, annotated according to whether the interaction ends with a ``late user-initiated hang-up or forward'' (LUHF), indicating user frustration and either call termination or escalation. Around 33\% of dialogues are labelled as breakdowns. To protect privacy, BETOLD encodes dialogues as sequences of intents and entities rather than raw text. We follow the official train–test splits and treat the provided labels as ground truth.

\subsection{Data Preprocessing and Experimental Setup}
We binarize labels on DBDC5 and BETOLD, merging Possible Breakdown with Breakdown, following~\cite{Ghassel2024Are, lin_semi-supervised_2022}. To accommodate privacy constraints, the BETOLD dataset utilizes intent and entity abstractions rather than raw utterances, potentially limiting the efficacy of specific prompting techniques. We parse these structured dialogues and present them in a text form like “System: Intent: X | Entities: Y” for each turn, thereby preserving the sequence and dialogue flow in anonymized form. Basic cleaning (removal of extraneous symbols, ensuring consistent turn indexing) is applied for both datasets. No additional data augmentation is used beyond what is provided in the datasets.

Our evaluation requires the LLM to provide three fields per response: justification, decision, and confidence score. Experimentally, the ordering of these requests influenced the model’s performance significantly. When the model was first prompted for a decision, subsequent justifications often appeared overly confident and less reflective. Conversely, requesting justifications first yielded more nuanced and deliberative decisions, enhancing overall output quality.

For AR and CL+AR, we restrict evaluation to 10\% subsamples of each dataset due to high token cost. Preliminary experiments indicate that a two-pass analogical pipeline (analogy generation followed by decision) performs comparably to single-pass prompting, aligning with existing literature~\cite{analogical}.

\subsection{Fine-tuning Parameters}
We fine-tune \texttt{Llama-3.1 8B} using Low-Rank Adaptation (LoRA, rank = 16)~\cite{hu2022lora}. We use AdamW with 8-bit parameters~\cite{adamw}, a learning rate of \(2 \times 10^{-4}\), linear decay, batch size 8, weight decay 0.01, and train for three epochs on a single NVIDIA A100 40GB GPU. We hypothesize that a carefully fine-tuned small LLM can provide competitive accuracy while remaining suitable for real-time monitoring. Our trained dialogue disruption monitor is released on HuggingFace\footnote{\url{https://huggingface.co/aghassel/dialogue\_disruption\_monitor}}. Because LoRA updates only a subset of parameters, this training regime substantially reduces compute and energy consumption compared to full-parameter fine-tuning.

\subsection{Evaluation Metrics}
We report accuracy and F1 scores for both Breakdown (B) and Non-Breakdown (NB) classes. Accuracy measures overall correctness; class-specific F1 scores capture performance on the minority Breakdown class without ignoring the majority NB class. Following DBDC5 challenge practice~\cite{ng_improving_2020, lin_semi-supervised_2022}, we report accuracy and per-class F1 on held-out evaluation sets (DBDC5 official evaluation and reserved BETOLD test splits).

Models are prompted to output JSON-formatted responses. We parse these outputs to extract the decision label. When outputs deviate from the expected format, we employ \texttt{Llama-3.3 70B} as an LLM-based judge to interpret the textual response and recover the intended classification.

On DBDC5 English and Japanese, we also assess overconfidence by computing MSE between the model’s verbalized confidence and the annotator-derived breakdown probabilities.

\subsection{Inference and Cost Considerations}
We evaluate a broad set of LLMs, summarized in Table \ref{tab:results}. Proprietary models include OpenAI’s \texttt{GPT-3.5 Turbo} and \texttt{GPT-4o}~\cite{openai_gpt-4_2023}, and Anthropic’s \texttt{Claude-3.5 Haiku} and \texttt{Claude-3.5 Sonnet}~\cite{ClaudeModelCard}. Open-source models include Mistral’s \texttt{Mixtral 8x7B} and \texttt{Mixtral 8x22B}~\cite{jiang_mixtral_2024}, DeepSeek’s \texttt{DeepSeek-R1}~\cite{deepseek}, and Meta’s \texttt{Llama-3.1 8B}, \texttt{Llama-3.3 70B}, and \texttt{Llama-3.1 405B}~\cite{llama}. Inference is run via the OpenRouter API\footnote{\url{https://openrouter.ai}} with a temperature of 0. To ensure comparability across models, we cap total tokens at 2{,}048 per query, which also reflects practical constraints for longer AR/CL+AR prompts.

Advanced prompting strategies such as AR and CL+AR require additional tokens, since models must generate analogous examples in context. We note this increased token usage as a trade-off: these strategies may improve reasoning at higher cost and latency.

\section{Results and Discussion}\label{res-and-disc}
We now analyze performance across BETOLD and the English and Japanese tracks of DBDC5. We compare prompting strategies, model families, and our fine-tuned Dialogue Disruption Monitor. Table~\ref{tab:results} reports accuracy, class-specific F1 scores, and calibration (MSE) where applicable.

\begin{table*}[ht!]
\centering
\scriptsize
\setlength{\tabcolsep}{6pt}
\renewcommand{\arraystretch}{0.9}
\caption{Results of proprietary and open-source models. Note: MSE values are scaled by 100 ($10^{-2}$). \textbf{Bold} is best, \underline{underlined} is second-best. Rows highlighted in grey indicate the BETOLD subset on specific prompting techniques.}
\label{tab:results}
\vspace{-1em}
\begin{tabular}{%
  >{\raggedright}m{1cm}   
  >{\raggedright}m{3cm}   
  l                         
  r r r                     
  r r r r                   
  r r r r                    
}
\toprule
\multirow{2}{*}{\textbf{Family}} & \multirow{2}{*}{\textbf{Model}} & \multirow{2}{*}{\textbf{Prompt}} 
& \multicolumn{3}{c}{\textbf{BETOLD}} 
& \multicolumn{4}{c}{\textbf{DBDC5 English}} 
& \multicolumn{4}{c}{\textbf{DBDC5 Japanese}} \\
\cmidrule(lr){4-6}\cmidrule(lr){7-10}\cmidrule(lr){11-14}
& &
& \textbf{Acc.} & \textbf{F1$_B$} & \textbf{F1$_{NB}$}
& \textbf{Acc.} & \textbf{F1$_B$} & \textbf{F1$_{NB}$} & \textbf{MSE}
& \textbf{Acc.} & \textbf{F1$_B$} & \textbf{F1$_{NB}$} & \textbf{MSE} \\
\midrule
\multirow{2}{*}{\shortstack{\textit{Previous}\\ \textit{SOTA}}}
& \texttt{BERT+SSMBA} & --
    & -- & -- & --
    & 73.9 & 78.2 & -- & --
    & -- & -- & -- & -- \\
\cmidrule(lr){2-14}
& \texttt{S2T2} & --
    & -- & -- & --
    & 77.9 & 82.4 & -- & --
    & 76.7 & 75.4 & -- & -- \\
\midrule
\multirow{14}{*}{Anthropic}
& \multirow{7}{*}{\texttt{Claude-3.5 Haiku}} & {ZS}
    & 74.4 & 55.0 & 82.1
    & 80.4 & 86.3 & 65.3 & 5.9
    & 74.6 & 76.0 & 73.0 & 7.3 \\
& 
  & CoT 
    & 73.9 & 52.6 & 82.0
    & 82.0 & 86.9 & 71.2 & 6.8
    & 77.1 & 76.5 & 77.6 & 8.2 \\
& 
  & 2S (Easy)
    & 74.0 & 64.3 & 79.5
    & 82.5 & 87.5 & 70.9 & 5.5
    & 68.9 & 73.3 & 62.9 & 10.6 \\
& 
  & 2S (Hard)
    & 74.1 & \underline{66.9} & 78.8
    & 82.5 & 87.4 & 70.9 & \underline{4.1}
    & 66.3 & 72.0 & 57.7 & 8.4 \\
& 
  & 4S
    & 74.1 & 67.0 & 78.7
    & 82.9 & 87.6 & 72.5 & 4.4
    & 67.0 & 72.3 & 59.2 & 10.2 \\
& 
  & AR
    & \cellcolor{gray!15}76.3 & \cellcolor{gray!15}\textbf{68.6} & \cellcolor{gray!15}81.0
    & 78.5 & 85.8 & 55.7 & --
    & 82.0 & 88.5 & 59.1 & -- \\
& 
  & CL+AR
    & \cellcolor{gray!15}77.0 & \cellcolor{gray!15}64.4 & \cellcolor{gray!15}83.1
    & 78.0 & 85.4 & 55.1 & --
    & 78.0 & 85.9 & 50.0 & -- \\
\cmidrule(lr){2-14}
& \multirow{7}{*}{\texttt{Claude-3.5 Sonnet}}
  & ZS
    & 75.1 & 54.9 & 82.8
    & 82.7 & 87.5 & 71.9 & 8.1
    & 81.3 & 81.2 & 81.4 & 8.2 \\
& 
  & CoT
    & 76.6 & 56.7 & \underline{84.0}
    & 82.5 & 87.3 & 71.6 & 7.8
    & 78.8 & 78.7 & 78.9 & 8.3 \\
& 
  & 2S (Easy)
    & 75.8 & 60.4 & 82.5
    & 83.5 & 88.2 & 72.3 & 7.1
    & 74.0 & 76.7 & 70.5 & 10.8 \\
& 
  & 2S (Hard)
    & 76.9 & 62.9 & 83.3
    & 81.2 & 86.9 & 66.7 & 7.5
    & 71.2 & 75.3 & 65.6 & 10.3 \\
& 
  & 4S
    & 76.7 & 64.4 & 82.7
    & 84.0 & 88.7 & 72.7 & 6.5
    & 71.4 & 75.4 & 65.9 & 10.7 \\
& 
  & AR
    & \cellcolor{gray!15}73.3 & \cellcolor{gray!15}55.0 & \cellcolor{gray!15}81.1
    & \textbf{85.5} & \textbf{89.8} & 74.8 & --
    & \underline{88.0} & \underline{91.7} & 78.6 & -- \\
& 
  & CL+AR
    & \cellcolor{gray!15}76.3 & \cellcolor{gray!15}63.6 & \cellcolor{gray!15}82.4
    & 83.5 & 88.5 & 70.8 & --
    & \textbf{89.0} & \textbf{92.4} & 80.0 & -- \\
\midrule
\multirow{14}{*}{OpenAI}
& \multirow{7}{*}{\texttt{GPT-3.5 Turbo}} & ZS
    & 41.2 & 51.4 & 25.8
    & 68.7 & 80.5 & 21.2 & 16.8
    & 50.8 & 64.4 & 20.4 & 25.2 \\
& 
  & CoT
    & 43.1 & 51.8 & 30.6
    & 67.9 & 77.9 & 41.9 & 17.8
    & 55.1 & 58.9 & 50.5 & 22.0 \\
& 
  & 2S (Easy)
    & 64.5 & 62.3 & 66.5
    & 67.2 & 76.6 & 44.9 & 14.7
    & 56.5 & 61.8 & 49.6 & 21.5 \\
& 
  & 2S (Hard)
    & 57.3 & 52.9 & 60.9
    & 67.0 & 74.7 & 52.8 & 8.6
    & 51.1 & 64.6 & 21.0 & 16.0 \\
& 
  & 4S
    & 47.5 & 55.0 & 37.1
    & 70.0 & 77.1 & 56.4 & 12.8
    & 57.1 & 57.7 & 56.4 & 20.5 \\
& 
  & AR
    & \cellcolor{gray!15}43.0 & \cellcolor{gray!15}52.2 & \cellcolor{gray!15}29.4
    & 71.5 & 81.9 & 32.9 & --
    & 71.0 & 82.2 & 21.6 & -- \\
& 
  & CL+AR
    & \cellcolor{gray!15}42.2 & \cellcolor{gray!15}51.9 & \cellcolor{gray!15}27.8
    & 68.0 & 79.2 & 30.4 & --
    & 72.0 & 81.8 & 39.1 & -- \\
\cmidrule(lr){2-14}
& \multirow{7}{*}{\texttt{GPT-4o}}
  & ZS & 74.1 & 47.8 & 82.7
    & 81.4 & 85.9 & 72.5 & 9.2
    & 79.2 & 76.7 & 81.2 & 9.8 \\
& 
  & CoT
    & 73.2 & 43.6 & 82.5
    & 82.3 & 86.5 & 74.3 & 9.1
    & 79.5 & 77.6 & 81.2 & 9.8 \\
& 
  & 2S (Easy)
    & 75.6 & 53.3 & 83.5
    & 82.7 & 86.7 & 75.2 & 7.2
    & 79.3 & 78.9 & 79.6 & 8.7 \\
& 
  & 2S (Hard)
    & \underline{77.4} & 63.0 & 83.7
    & 83.5 & 87.3 & 76.3 & 5.1
    & 80.2 & 77.7 & 82.1 & 6.4 \\
& 
  & 4S
    & \textbf{77.7} & 62.8 & \textbf{84.1}
    & 82.0 & 86.1 & 74.6 & 5.8
    & 79.8 & 79.2 & 80.4 & 7.5 \\
& 
  & AR
    & \cellcolor{gray!15}70.4 & \cellcolor{gray!15}42.9 & \cellcolor{gray!15}80.0
    & 80.5 & 86.3 & 66.1 & --
    & 87.0 & 91.0 & 76.4 & -- \\
& 
  & CL+AR
    & \cellcolor{gray!15}70.4 & \cellcolor{gray!15}44.4 & \cellcolor{gray!15}79.8
    & 81.5 & 86.4 & 70.9 & --
    & 85.0 & 89.4 & 74.6 & -- \\
\midrule
\multirow{22}{*}{\centering Meta}%
& \multirow{7}{*}{\texttt{Llama-3.1 8B}}
  & ZS
    & 60.2 & 59.9 & 60.6
    & 73.2 & 80.4 & 57.8 & 9.1
    & 65.0 & 66.9 & 62.9 & 12.3 \\
& 
  & CoT
    & 56.3 & 56.3 & 56.4
    & 73.4 & 81.2 & 54.6 & 12.1
    & 60.6 & 66.0 & 53.2 & 12.5 \\
& 
  & 2S (Easy)
    & 69.2 & 44.0 & 78.7
    & 75.6 & 82.6 & 59.6 & 8.9
    & 59.7 & 68.2 & 45.1 & 16.7 \\
& 
  & 2S (Hard)
    & 68.3 & 53.4 & 76.0
    & 73.7 & 81.9 & 52.5 & 8.3
    & 59.7 & 66.4 & 49.8 & 12.6 \\
& 
  & 4S
    & 71.6 & 50.9 & 80.0
    & 76.4 & 83.3 & 60.1 & 6.9
    & 59.9 & 66.7 & 49.7 & 11.7 \\
& 
  & AR
    & \cellcolor{gray!15}65.9 & \cellcolor{gray!15}60.3 & \cellcolor{gray!15}70.1
    & 64.0 & 75.0 & 35.7 & --
    & 69.0 & 78.9 & 41.5 & -- \\
& 
  & CL+AR
    & \cellcolor{gray!15}60.0 & \cellcolor{gray!15}60.4 & \cellcolor{gray!15}67.1
    & 66.5 & 78.3 & 26.4 & --
    & 67.0 & 79.2 & 19.5 & -- \\
\cmidrule(lr){2-14}
& \multirow{7}{*}{\texttt{Llama-3.3 70B}}
  & ZS
    & 72.7 & 38.5 & 82.4
    & 83.0 & 87.4 & 74.2 & 6.2
    & 77.9 & 76.8 & 78.9 & 7.8 \\
& 
  & CoT
    & 74.0 & 49.7 & 82.5
    & 81.7 & 86.2 & 73.0 & 6.0
    & 76.3 & 76.0 & 76.7 & 7.7 \\
& 
  & 2S (Easy)
    & 73.1 & 36.4 & 82.9
    & 81.7 & 85.5 & 75.3 & 6.2
    & 76.9 & 77.9 & 75.8 & 8.7 \\
& 
  & 2S (Hard)
    & 74.1 & 43.0 & 83.3
    & 82.6 & 86.2 & \underline{76.4} & \textbf{4.0}
    & 77.8 & 77.1 & 78.4 & 5.8 \\
& 
  & 4S
    & 73.2 & 40.4 & 82.7
    & 81.9 & 85.5 & 75.8 & 4.7
    & 78.7 & 78.4 & 79.0 & 6.7 \\
& 
  & AR
    & \cellcolor{gray!15}70.4 & \cellcolor{gray!15}28.6 & \cellcolor{gray!15}81.3
    & 84.5 & 88.0 & 78.0 & --
    & 84.0 & 88.1 & 75.8 & -- \\
& 
  & CL+AR
    & \cellcolor{gray!15}72.6 & \cellcolor{gray!15}43.1 & \cellcolor{gray!15}82.0
    & \textbf{85.5} & \underline{89.5} & \textbf{76.8} & --
    & 77.0 & 83.1 & 67.6 & -- \\
\cmidrule(lr){2-14}
& \multirow{7}{*}{\texttt{Llama-3.1 405B}}
  & ZS
    & 71.2 & 30.1 & 81.8
    & 81.5 & 86.3 & 71.4 & 6.2
    & 78.7 & 77.0 & 80.2 & 6.1 \\
& 
  & CoT
    & 72.1 & 35.1 & 82.3
    & 80.7 & 85.5 & 71.2 & 6.3
    & 79.5 & 77.0 & 81.6 & 6.1 \\
& 
  & 2S (Easy)
    & 72.6 & 45.5 & 81.7
    & 81.0 & 84.6 & 75.3 & 6.4
    & 80.7 & 78.7 & \underline{82.4} & 7.1 \\
& 
  & 2S (Hard)
    & 74.0 & 45.7 & 82.9
    & 79.4 & 82.9 & 74.1 & 4.3
    & 79.4 & 75.5 & 82.2 & \textbf{4.6} \\
& 
  & 4S
    & 75.2 & 51.0 & 83.4
    & 79.6 & 82.9 & 74.8 & 5.6
    & 81.0 & 78.4 & \textbf{83.1} & \underline{5.8} \\
& 
  & AR
    & \cellcolor{gray!15}65.2 & \cellcolor{gray!15}29.9 & \cellcolor{gray!15}76.8
    & 77.0 & 82.8 & 65.2 & --
    & 72.0 & 78.1 & 61.1 & -- \\
& 
  & CL+AR
    & \cellcolor{gray!15}58.5 & \cellcolor{gray!15}22.2 & \cellcolor{gray!15}71.4
    & 79.0 & 84.6 & 67.2 & --
    & 82.0 & 87.1 & 70.0 & -- \\
\cmidrule(lr){2-14}
& \multicolumn{1}{l}{{\texttt{Disruption Monitor}}}
  & -
    & 67.2 & 59.7 & 72.3
    & 81.5 & 86.2 & 72.0 & 4.9
    & 67.9 & 68.8 & 66.9 & 8.8 \\
\midrule
\multirow{14}{*}{Mistral}
& \multirow{7}{*}{\texttt{Mixtral 8x7B}} & ZS
    & 70.5 & 42.4 & 80.2
    & 58.4 & 56.1 & 60.5 & 11.9
    & 62.5 & 37.4 & 73.2 & 13.5 \\
& 
  & CoT
    & 67.0 & 21.5 & 79.1
    & 57.9 & 54.6 & 60.8 & 13.2
    & 64.4 & 40.4 & 74.6 & 13.9 \\
& 
  & 2S (Easy)
    & 69.7 & 30.0 & 80.7
    & 64.9 & 65.6 & 64.2 & 10.3
    & 67.4 & 54.3 & 74.7 & 11.9 \\
& 
  & 2S (Hard)
    & 71.5 & 50.8 & 79.9
    & 68.0 & 69.5 & 66.4 & 6.3
    & 66.8 & 52.2 & 74.5 & 10.1 \\
& 
  & 4S
    & 71.2 & 43.0 & 80.8
    & 68.1 & 69.9 & 66.0 & 8.3
    & 68.6 & 55.0 & 75.9 & 11.4 \\
& 
  & AR
    & \cellcolor{gray!15}65.2 & \cellcolor{gray!15}43.6 & \cellcolor{gray!15}75.1
    & 58.5 & 66.9 & 44.3 & --
    & 60.0 & 69.7 & 41.2 & -- \\
& 
  & CL+AR
    & \cellcolor{gray!15}65.9 & \cellcolor{gray!15}44.4 & \cellcolor{gray!15}79.8
    & 62.5 & 70.1 & 49.7 & --
    & 62.0 & 68.9 & 51.3 & -- \\
\cmidrule(lr){2-14}
& \multirow{7}{*}{\texttt{Mixtral 8x22B}}
  & ZS
    & 70.7 & 28.5 & 81.6
    & 81.8 & 85.9 & 74.7 & 8.9
    & 74.3 & 64.8 & 79.8 & 14.3 \\
& 
  & CoT
    & 70.8 & 33.4 & 81.3
    & 80.9 & 85.4 & 72.5 & 9.0
    & 73.7 & 64.0 & 79.3 & 14.7 \\
& 
  & 2S (Easy)
    & 73.8 & 55.8 & 81.4
    & 83.5 & 87.4 & 76.0 & 6.7
    & 78.9 & 75.8 & 81.3 & 9.9 \\
& 
  & 2S (Hard)
    & 73.2 & 61.7 & 79.4
    & 81.1 & 84.9 & 74.5 & 4.2
    & 75.5 & 69.9 & 79.4 & 6.6 \\
& 
  & 4S
    & 75.2 & 60.4 & 82.0
    & 81.9 & 85.7 & 75.5 & 5.9
    & 76.2 & 71.0 & 79.8 & 9.5 \\
& 
  & AR
    & \cellcolor{gray!15}68.1 & \cellcolor{gray!15}31.7 & \cellcolor{gray!15}79.2
    & 81.0 & 86.8 & 66.1 & --
    & 79.0 & 84.2 & 68.7 & -- \\
& 
  & CL+AR
    & \cellcolor{gray!15}63.7 & \cellcolor{gray!15}47.3 & \cellcolor{gray!15}72.3
    & 75.5 & 84.0 & 47.3 & --
    & 83.0 & 88.3 & 69.1 & -- \\
\midrule
\multirow{7}{*}{DeepSeek}
& \multirow{7}{*}{\texttt{DeepSeek-R1}}
  & ZS
    & 73.8 & 55.9 & 81.3
    & 81.1 & 86.4 & 69.1 & 6.7
    & 74.5 & 76.7 & 71.8 & 8.8 \\
& 
  & CoT
    & 74.9 & 57.5 & 82.2
    & 80.4 & 86.0 & 67.6 & 6.6
    & 76.6 & 78.1 & 74.9 & 7.6 \\
& 
  & 2S (Easy)
    & 75.8 & 57.8 & 83.0
    & 82.3 & 87.0 & 72.5 & 6.4
    & 72.7 & 75.6 & 69.0 & 10.1 \\
& 
  & 2S (Hard)
    & 76.4 & 65.4 & 82.1
    & 82.0 & 86.9 & 71.4 & 4.4
    & 72.5 & 76.1 & 67.7 & 6.9 \\
& 
  & 4S
    & 75.8 & 61.7 & 82.3
    & 83.0 & 87.2 & 74.5 & 4.5
    & 74.7 & 77.1 & 71.8 & 7.7 \\
& 
  & AR
    & \cellcolor{gray!15}71.1 & \cellcolor{gray!15}60.6 & \cellcolor{gray!15}77.2
    & 80.0 & 86.2 & 63.6 & --
    & 85.0 & 90.2 & 68.1 & -- \\
& 
  & CL+AR
    & \cellcolor{gray!15}75.6 & \cellcolor{gray!15}66.7 & \cellcolor{gray!15}80.7
    & 80.0 & 86.5 & 61.5 & --
    & 87.0 & 91.4 & 73.5 & -- \\
\bottomrule
\end{tabular}
\end{table*}

\subsection{Main Findings}

\noindent \textbf{State-of-the-Art Results.}
On DBDC5 English, multiple modern LLMs surpass the prior best accuracy of 77.9\% from \texttt{S2T2}. \texttt{Claude-3.5 Sonnet} and \texttt{Llama-3.3 70B} both reach 85.5\% accuracy under AR and CL+AR prompting, respectively, achieving state-of-the-art performance. On DBDC5 Japanese, \texttt{Claude-3.5 Sonnet} with CL+AR attains 89.0\% accuracy, slightly ahead of the strongest open-source model, \texttt{DeepSeek-R1}, at 87.0\%. Overall, open-source models are now within 1–3 points of leading closed-source systems on these benchmarks.

\noindent \textbf{Closed-Source Frontier Models}.
Claude-3.5 (\texttt{Haiku}, \texttt{Sonnet}) demonstrate strong classification consistency across both DBDC5 tracks, ranging from 74\% to 89\% on Japanese and 78\% to 85\% on English. In particular, \texttt{Sonnet} combined with AR or CL+AR prompts yields top accuracies, for instance, 85.5\% on English and 89.0\% on Japanese. GPT-4o likewise competes closely, achieving up to 77.7\% on BETOLD via 4S prompting and 83.5\% on DBDC5 English using a more challenging 2S-Hard strategy. While marginally behind \texttt{Claude-3.5 Sonnet} on Japanese, GPT-4o’s performance remains robust, although it exhibits greater sensitivity to variations in prompt style.

In contrast, \texttt{GPT-3.5 Turbo} underperforms substantially on BETOLD, with accuracy ranging from 41\% to 64.5\%. It tends to misclassify borderline “near-breakdown” utterances or produce imbalanced predictions. Error analysis suggests that \texttt{GPT-3.5 Turbo} is more sensitive to how examples are presented; certain prompt structures lead to skewed confidence or confusion in distinguishing near-breakdown from non-breakdown scenarios.

\noindent \textbf{Open-Source Models.}
Larger open-source models (\texttt{Llama-3.3 70B}, \texttt{Llama-3.1 405B}, \texttt{Mixtral 8x22B}, \texttt{DeepSeek-R1}) match or exceed closed-source baselines on DBDC5 English (80\%-85\%) and demonstrate competitive performance on DBDC5 Japanese. However, on BETOLD, performance variability is pronounced, ranging from 68.3\% (\texttt{Llama-3.1 8B}, 2S-Hard) to 75.8\% (\texttt{DeepSeek-R1}, 2S-Easy), reflecting difficulties generalizing from natural dialogues to abstract intent representations. \texttt{Llama-3.1 405B}, despite its size, does not consistently surpass its 70B counterpart. Its best English-track accuracy hovers near 79\%–81\%, indicating marginal variances to prompting techniques. 

\noindent \textbf{Dialogue Disruption Monitor.}
Despite its size, our fine-tuned \texttt{Llama-3.1 8B} monitor achieves 81.5\% accuracy on DBDC5 English, exceeding several larger models and approaching \texttt{Llama-3.3 70B}. On DBDC5 Japanese, it reaches 67.9\% accuracy with balanced F1 scores, reflecting limited Japanese coverage in pretraining. On BETOLD, the monitor achieves 67.2\% accuracy, a 7-point absolute improvement over the zero-shot \texttt{Llama-3.1 8B} baseline (60.2\%). This demonstrates cross-dataset transfer of breakdown detection behaviour. On DBDC5 English, the monitor’s calibration (MSE = 4.9) is competitive with much larger models. To improve Japanese performance further, future work could explore models such as \texttt{Llama-3.1 Swallow 8B}, which are pre-trained on large-scale Japanese corpora~\cite{swallow:COLM2024}.

\subsection{Impact of Prompting Strategies}
\noindent \textbf{Few-Shot Prompting. }
Few-shot prompting consistently outperforms ZS and CoT approaches across datasets. For instance, \texttt{DeepSeek-R1} rises from 81.1\% (ZS) to 83.0\% (4S). The “Hard” exemplars comprised of borderline dialogues, also yield stronger calibration (lower MSE). 2S-Hard achieves substantial calibration improvements, with \texttt{Llama-3.3 70B} reaching an MSE of 4.0 on DBDC5 English, the lowest among all models. Similarly, \texttt{GPT-4o}'s accuracy improves from 81.4\% (ZS) to 83.5\% (2S-Hard), affirming the effectiveness of providing challenging examples to refine model uncertainty estimations more effectively than simpler or even more examples.

\noindent \textbf{Limitations of Chain-of-Thought. }
CoT yields mixed results, slightly improving borderline-case identification. For instance, \texttt{GPT-4o} on DBDC5 English F1(B) improves from 85.9\% to 86.5\%. However, this occasionally degrades performance on structured datasets like BETOLD, indicating that an optimal reasoning complexity is dataset-dependent. This highlights the importance of tailoring reasoning complexity to the task: dialogues with short turns or highly structured, domain-specific content may be more effectively processed using concise prompts rather than elaborate `think step-by-step' sequences.

\noindent \textbf{AR and CL+AR on BETOLD. } 
While AR and CL+AR techniques improved performance on natural dialogue datasets (DBDC5 tracks), their effectiveness on BETOLD was mixed, with larger models showing degradation while some smaller models benefited (highlighted in grey in Table~\ref{tab:results}). This degradation may stem from BETOLD's representation of dialogues through structured intents and entities rather than natural language utterances. This mismatch undermines generalization, as models were predominantly trained on conventional dialogue data.
Error analysis revealed frequent instruction-following failures. Models either generated no analogous examples (instead directly solving the original dialogue) or produced overly brief analogies with weak alignment to the target dialogue. Even \texttt{GPT-4o} struggled with reliable compliance (Figure~\ref{fig:error:gpt-4-analog-betold}). BETOLD's longer dialogues (20–30 turns vs. 10–20 turns in DBDC5) exacerbate these issues. Three analogous examples plus the original conversation frequently exhausted the token budget (Figure~\ref{fig:betold:cl+ar-prompt}). Smaller variants (\texttt{Llama-3.1 8B}, \texttt{Mixtral 8x7B}, \texttt{Claude-3.5 Haiku}) showed poor analogy generation, likely due to their reduced capacity and limited training.

\subsection{Calibration and Confidence Analysis}
Low MSE indicates good alignment between model confidence and human agreement. On DBDC5 English, \texttt{Llama-3.3 70B} with 2S-Hard achieves the best calibration (MSE = 4.0), while \texttt{Llama-3.1 405B} with 2S-Hard obtains the best calibration on Japanese (MSE = 4.6). Among proprietary models, \texttt{Claude-3.5 Haiku} reaches MSE = 4.1 on English under 2S-Hard. Our 8B monitor’s MSE of 4.9 demonstrates that moderate-scale, task-specific fine-tuning can yield calibration quality close to much larger models.

\subsection{Sensitivity Analysis}
We analyze how the escalation threshold $T$ trades off safety and cost. For each dataset, we vary $T$ (minimum confidence required to accept a Non-Breakdown decision without escalation) and report safety (recall for the Breakdown class) and cost (escalation rate, i.e., the fraction of turns escalated).

\noindent \textbf{Decisive Regime ($T \le 0.5$).}
Across DBDC5 English, DBDC5 Japanese, and BETOLD, the curves are largely flat for $T \in [0, 0.5]$, indicating that the monitor is confident when predicting Non-Breakdown. In this regime the system behaves like a near-binary classifier with strong safety at low manual tuning effort. For example, on DBDC5 English, we obtain $\approx$88\% recall while escalating only $\approx$70\% of turns, about a 30\% reduction in compute relative to full escalation.

\noindent \textbf{Calibration Regime ($T > 0.5$).}
Beyond $T=0.5$, the monitor begins escalating low-confidence non-breakdown predictions, yielding coupled increases in safety and cost. This reflects an effective uncertainty safety net: raising $T$ recovers borderline breakdowns that would otherwise be missed. On the BETOLD set, increasing $T$ from 0.5 to 0.8 boosts safety from roughly 70\% to $>90\%$, at the expense of higher escalation.

\noindent \textbf{Operational Guidance.}
The architecture exposes a simple knob for budget--safety trade-offs. A default of $T=0.5$ yields substantial efficiency gains for standard deployments. For high-stakes domains where missed breakdowns are unacceptable, \(T\) can be increased to push breakdown recall closer to 100\% while still avoiding full reliance on the largest model.

\begin{figure*}[t]
    \centering
    \subfloat[\textbf{DBDC5 English}: High baseline safety with $\sim$30\% fewer escalations.]{%
        \includegraphics[width=0.32\textwidth]{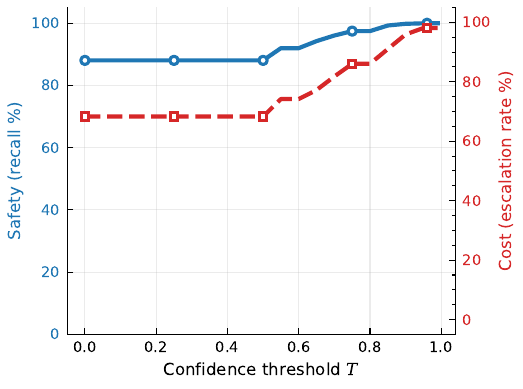}%
        \label{fig:sens_en}%
    }%
    \hfill
    \subfloat[\textbf{DBDC5 Japanese}: Similar profile, indicating multilingual stability.]{%
        \includegraphics[width=0.32\textwidth]{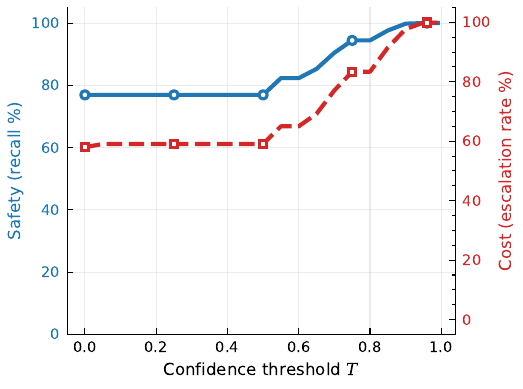}%
        \label{fig:sens_ja}%
    }%
    \hfill
    \subfloat[\textbf{BETOLD}: Lower due to domain shift; tuning $T$ pushes safety $>$90\%.]{%
        \includegraphics[width=0.32\textwidth]{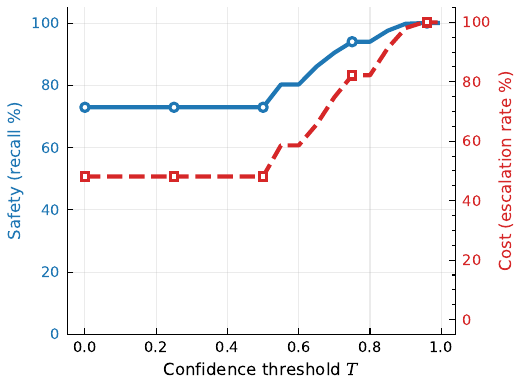}%
        \label{fig:sens_betold}%
    }
    \caption{\textbf{Sensitivity to the escalation threshold.} As $T$ increases, safety (breakdown recall) and cost (escalation rate) trace a Pareto-like frontier. The flat segment for $T \le 0.5$ reflects decisive Non-Breakdown predictions; the rising segment for $T>0.5$ shows that computation can be traded for additional safety.}
    \label{fig:sensitivity}
\end{figure*}

\subsection{End-to-End Repair Capabilities}
To assess closed-loop remediation, we sampled 100 correctly detected breakdowns from the DBDC5 English test set. We passed the dialogue history and the monitor's justification ($\hat{j}_i$) to the Superior Model (\texttt{Claude-3.5 Sonnet}) with the instruction: \textit{``The following response was flagged as a breakdown because: [Justification]. Please rewrite the response to be coherent and consistent with the dialogue history.''}

An LLM-as-a-judge~\cite{Gu2024A} evaluation confirmed that conditioning on these justifications resolved 97\% of sampled breakdowns. In contrast, a baseline prompted with the generic instruction \textit{``Please rewrite the response to be coherent and consistent with the dialogue history''} (without the monitor's explanation) achieved a lower resolution rate of 92\%. Qualitative analysis indicates that the superior model leverages specific details in the monitor's flag, such as noting a direct contradiction with a previous turn, to generate targeted corrections rather than simply attempting a generic regeneration.

\subsection{Resource Efficiency and Sustainability}
Resource efficiency is a crucial factor influencing the practical deployment of language models. Advanced models such as \texttt{Claude-3.5 Sonnet} (400B parameters), \texttt{DeepSeek-R1} (671B parameters), and \texttt{GPT-4} (estimated 1.7 trillion parameters\cite{epochFrontierLanguage}) demand substantial monetary costs, experience network-induced latency, and have a higher carbon footprint per query due to their considerable size and reliance on data-centre computation. For example, based on cloud-provider pricing from Amazon Web Services (AWS), querying \texttt{Llama-3.1 70B} is approximately 3.3 times more expensive than its 8B-parameter counterpart, while \texttt{Llama-3.1 405B} incurs a cost approximately 10.3 times higher~\cite{amazonbedrockpricing}. Note, this also does not consider the latency, which in practice also increases substantially with model size, network routing complexity, and prompt length. Additionally, using longer prompts (such as AR or CL+AR) significantly escalates costs due to the increased token usage per query.

In contrast, our fine-tuned \texttt{Llama-3.1 8B} monitor runs efficiently on a single A100 GPU, with latency under half a second per dialogue turn, making it suitable for per-turn monitoring. In our cost analysis (Appendix~\ref{sec:bedrock-cost-example}), we estimate that a hierarchical deployment combining a 70B assistant, 405B escalations on 10\% of turns, and an 8B monitor achieves a 54\% cost reduction relative to running the entire dialogue on \texttt{Llama-3.1 405B}. Given that large deployments such as ChatGPT may consume on the order of 1{,}058.5 GWh annually~\cite{bestbrokersAIsPower}, invoking large models only when necessary can meaningfully reduce cost and environmental impact.

\subsection{Relevance of Breakdown Datasets to Modern LLMs}
Although DBDC5 was developed for chat-oriented dialogue systems rather than modern LLM deployments, its breakdown labels and error categories continue to capture core failure modes that remain prevalent in current deployments \cite{Higashinaka2020DBDC5}. DBDC5 explicitly models phenomena such as contradiction, wrong or missing information, ignored questions, and topic-transition errors, which map closely onto “intrinsic hallucinations” (conflicts with source context) and broader faithfulness errors in recent hallucination taxonomies~\cite{Huang2025Taxonomy,Higashinaka2020DBDC5}. Consequently, detecting these breakdown types remains a useful proxy for assessing LLM reliability on fundamental conversational errors, even as underlying architectures and training regimes evolve.

\section{Conclusion}\label{con-future-work}
Robustness in conversational AI is advancing along several dimensions: larger and better-trained LLMs, more targeted benchmarks such as DBDC5 and BETOLD, and reasoning techniques such as chain-of-thought, analogical prompting, and curriculum learning. Our work combines these elements into a practical framework for dialogue breakdown management. Empirically, we show that modern LLMs, both proprietary and open-source, achieve state-of-the-art performance on dialogue breakdown detection benchmarks. However, a gap remains to human-level performance, especially in handling complex and ambiguous conversations. Advanced prompting (e.g., CoT, AR) can yield further gains for some high-capacity models, but improvements are not uniform across datasets. Short, well-chosen exemplars often strike a better balance between performance and token usage. We find that eliciting both justifications and numeric confidence can reduce overconfidence and improve calibration, particularly when the model is prompted to justify before deciding. For high-volume, high-stakes applications, coupling a fast, fine-tuned monitor with on-demand escalation to a frontier LLM provides a practical path to reliability and sustainability. While closed-source models currently lead in absolute accuracy, careful tuning of small open-source models narrows the gap. Future work should explore more fine-grained breakdown taxonomies, richer multimodal settings, and tighter integration of monitoring signals into policy-level decisions, with the goal of building interpretable, and resource-aware conversational systems at scale.

\section{Limitations}
Although our approach performs competitively on both English and Japanese DBDC5 benchmarks, the smaller Llama-3.1 8B model shows limited Japanese coverage, suggesting that stronger multilingual pretraining or dedicated Japanese adaptation (e.g., using Swallow~\cite{swallow:COLM2024}) is needed. Moreover, real-world dialogues involving code-switching, adversarial inputs, or highly specialized domains may demand additional adaptations beyond the benchmarks we evaluate.
Second, prompt engineering remains model- and dataset-specific, and methods such as AR and CL+AR can be brittle and expensive. Moreover, LLMs remain partially opaque: even with chain-of-thought or analogical prompting, internal reasoning processes are not fully observable. Escalating to larger models introduces latency and energy overhead, which may be problematic in ultra-low-latency or resource-constrained settings.
Finally, our binary breakdown labels elide nuanced distinctions between mild and severe breakdowns and do not explicitly capture repair quality or user satisfaction. More expressive annotation schemes and joint modelling of detection and repair could enable finer-grained control in deployment. We leave these directions to future work.

\bibliographystyle{IEEEtran}
\bibliography{references}

\newpage
\appendix
\section{Cost-Savings Example on AWS Bedrock}
\label{sec:bedrock-cost-example}

This section quantifies the benefit of selective model escalation.  
We consider a 15-turn dialogue that normally runs on \texttt{Llama-3.1 70B}.  
When a real-time monitor detects a potential breakdown, the request is
re-issued to the larger \texttt{Llama-3.1 405B}. Empirically, such
escalations are required on roughly \(10\%\) of turns.

Table \ref{tab:bedrock-pricing} lists on-demand prices\footnote{\url{https://aws.amazon.com/bedrock/pricing/}}
for Meta’s Llama-3.1 family. Although Bedrock bills input and output tokens 
separately, the rates per 1k tokens are identical for these models; 
we therefore calculate costs using the total token count and the applicable rate.

\begin{table}[h]
  \setlength{\tabcolsep}{4pt}
  \centering
  \caption{Representative AWS Bedrock prices for Llama-3.1 (May 2025).}
  \label{tab:bedrock-pricing}
  \begin{tabular}{lcc}
    \toprule
    \textbf{Model} & \textbf{Input (per 1k)} & \textbf{Output (per 1k)} \\
    \midrule
    \texttt{Llama-3.1 8B}   & \$0.00022 & \$0.00022 \\
    \texttt{Llama-3.1 70B}  & \$0.00072 & \$0.00072 \\
    \texttt{Llama-3.1 405B} & \$0.00240 & \$0.00240 \\
    \bottomrule
  \end{tabular}
\end{table}

\subsection{Token Budget for 15 Turns}

Each turn contributes \(\approx\!40\) tokens in the user prompt and
another \(40\) tokens in the reply, for \(80\) new tokens per turn. Because
the entire history is sent at every step, turn~\(i\) carries \(80\,i\)
tokens.  Over 15 turns:

\[
\text{Total tokens} \;=\; 80 \sum_{i=1}^{15} i
      \;=\; 80 \times \frac{15 \times 16}{2}
      \;=\; 9\,600\text{ tokens.}
\]

\subsection{Baseline Costs}

\begin{itemize}
\item \textbf{Always 405B}  
      \(\;9.6 \times \$0.00240 = \$0.02304\)

\item \textbf{Always 70B}  
      \(\;9.6 \times \$0.00072 = \$0.00691\)
\end{itemize}

\subsection{Selective Escalation (70B + 10\% 405B)}

\[
\begin{aligned}
\text{70B: } & 0.9 \times 9\,600 = 8\,640 \text{ tokens}
               &&\Rightarrow 8.64 \times \$0.00072 \approx \$0.00622,\\
\text{405B: } & 0.1 \times 9\,600 = 960 \text{ tokens}
               &&\Rightarrow 0.96 \times \$0.00240 \approx \$0.00230,\\
\textbf{Total} & &&\approx \$0.00852.
\end{aligned}
\]

\subsection{Adding a Lightweight Monitor}

\noindent The monitor itself runs on \texttt{Llama-3.1 8B} and processes every
turn:

\[
9\,600 \text{ tokens} \;\times\; \$0.00022
      \;\approx\; \$0.00211.
\]

\medskip
\noindent Putting it all together:

\[
\$0.00211 \;(\text{monitor}) \;+\; \$0.00852 \;(\text{dialogue})
       \;=\; \$0.01063.
\]

Selective escalation with monitoring cuts cost by
\(\displaystyle 1 - \tfrac{0.01063}{0.02304} \approx 54\%\)
relative to running the entire conversation on \texttt{Llama-3.1 405B},
yet preserves the option to leverage the larger model when necessary.

\section{Additional Figures}
\onecolumn
\begin{figure}
    \centering
    \includegraphics[width=0.8\linewidth]{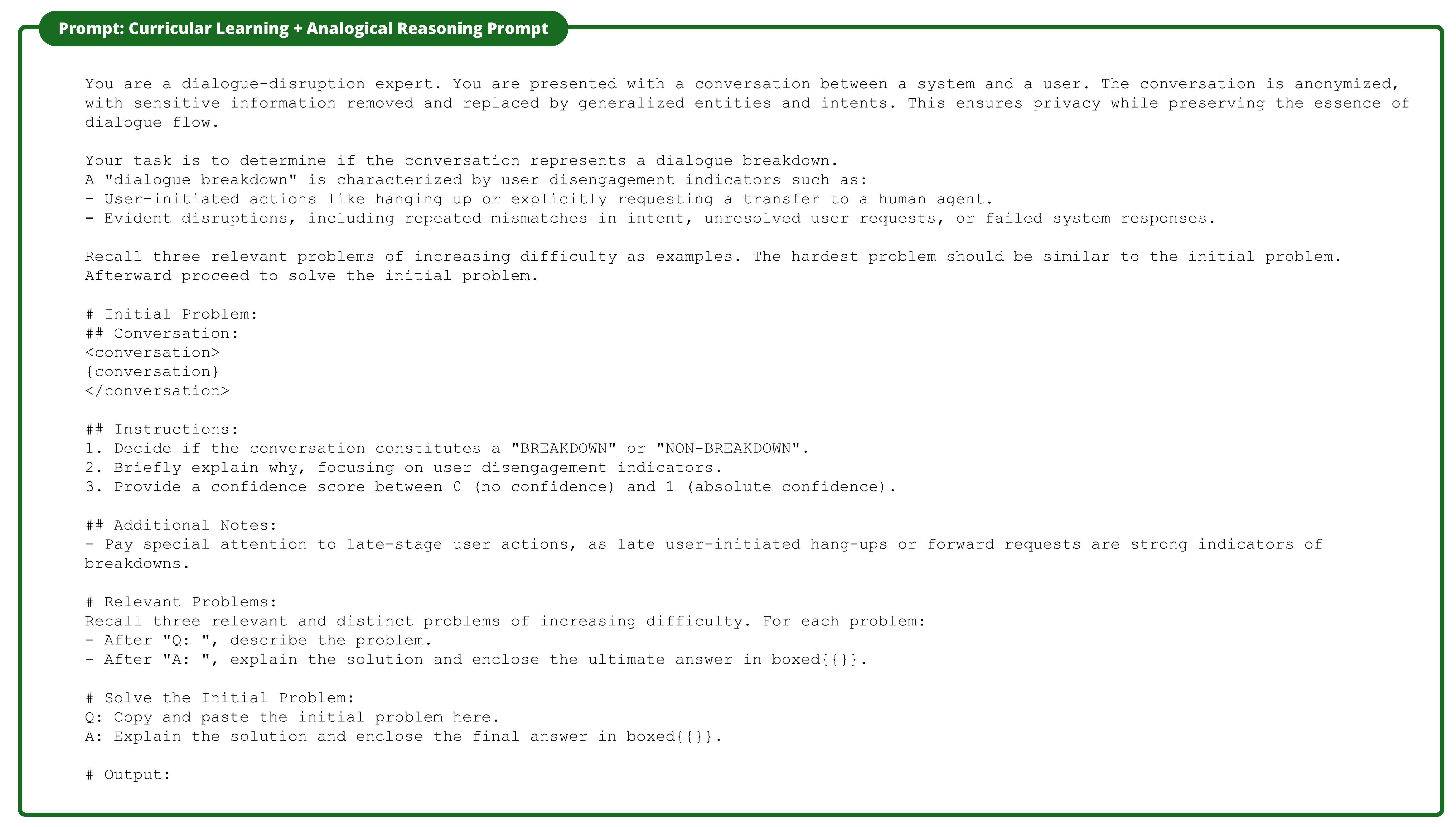}
    \caption{BETOLD: CL+AR Prompt}
    \label{fig:betold:cl+ar-prompt}
\end{figure}

\begin{figure}
    \centering
    \includegraphics[width=0.8\textwidth]{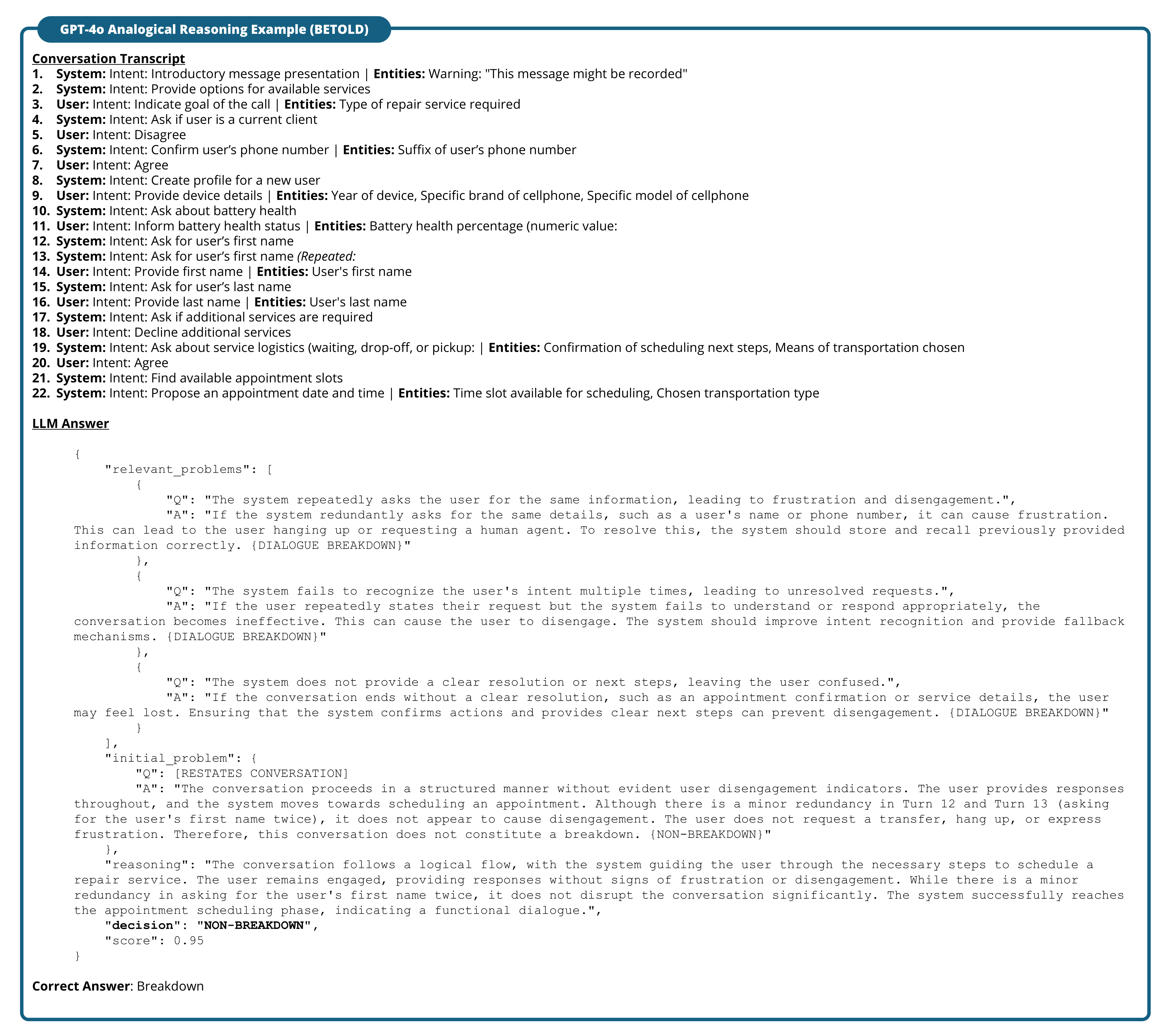}
    \caption{Error Analysis: GPT-4 Analogical Reasoning Example on BETOLD}
    \label{fig:error:gpt-4-analog-betold}
\end{figure}
\twocolumn

\end{document}